\documentclass[letterpaper]{article} 
\usepackage{aaai25}  
\usepackage{times}  
\usepackage{helvet}  
\usepackage{courier}  
\usepackage[hyphens]{url}  
\usepackage{graphicx} 
\usepackage{natbib}  
\usepackage{caption} 
\usepackage{algorithm}
\usepackage{algorithmic}
\usepackage{amsmath}
\usepackage{multirow}
\usepackage{color}

\usepackage{placeins}
\usepackage{amsfonts}
\usepackage{newfloat}
\usepackage{listings}
\DeclareCaptionStyle{ruled}{labelfont=normalfont,labelsep=colon,strut=off} 
\floatstyle{ruled}
\newfloat{listing}{tb}{lst}{}
\floatname{listing}{Listing}
\title{SG-MIM: Structured Knowledge Guided Efficient Pre-training\\for Dense Prediction}
\author {
    Sumin Son\textsuperscript{\rm 1},
    Hyesong Choi\textsuperscript{\rm 1},
    Dongbo Min\textsuperscript{\rm 1}
}
\affiliations {
    \textsuperscript{\rm 1}Ewha W. University\\
    sumin.son@ewha.ac.kr,
    hyesong@ewha.ac.kr,
    dbmin@ewha.ac.kr
}

\usepackage{bibentry}

\begin{document}

\maketitle

\begin{abstract}
Masked Image Modeling (MIM) techniques have redefined the landscape of computer vision, enabling pre-trained models to achieve exceptional performance across a broad spectrum of tasks. Despite their success, the full potential of MIM-based methods in dense prediction tasks, particularly in depth estimation, remains untapped. Existing MIM approaches primarily rely on single-image inputs, which makes it challenging to capture the crucial structured information, leading to suboptimal performance in tasks requiring fine-grained feature representation. To address these limitations, we propose SG-MIM, a novel Structured knowledge Guided Masked Image Modeling framework designed to enhance dense prediction tasks by utilizing structured knowledge alongside images. SG-MIM employs a lightweight relational guidance framework, allowing it to guide structured knowledge individually at the feature level rather than naively combining at the pixel level within the same architecture, as is common in traditional multi-modal pre-training methods. This approach enables the model to efficiently capture essential information while minimizing discrepancies between pre-training and downstream tasks. Furthermore, SG-MIM employs a selective masking strategy to incorporate structured knowledge, maximizing the synergy between general representation learning and structured knowledge-specific learning. Our method requires no additional annotations, making it a versatile and efficient solution for a wide range of applications. Our evaluations on the KITTI, NYU-v2, and ADE20k datasets demonstrate SG-MIM's superiority in monocular depth estimation and semantic segmentation.
\end{abstract}

%

\section{Introduction}

In the field of computer vision, pre-training with supervised classification on ImageNet~\cite{deng2009imagenet} has long been the gold standard, consistently demonstrating its unmatched effectiveness across a broad spectrum of visual tasks, particularly in tasks related to semantic understanding, such as image classification~\cite{kornblith2019better,dosovitskiy2020image,liu2021swin}, semantic segmentation~\cite{long2015fully,wang2018non,cheng2022masked}, and object detection~\cite{he2017mask,redmon2016you,carion2020end}. 
Building on this foundation, self-supervised pre-training methods—most notably 'Masked Image Modeling'~\cite{he2022masked,xie2022simmim,choi2024emerging,choi2024salience}, where the model learns to reconstruct randomly masked portions of an image—have become the leading approach, achieving superior performance across a range of downstream tasks.
The success of Masked Image Modeling (MIM) can be attributed significantly to the role of locality inductive bias~\cite{xie2023revealing}. Contrasted with supervised pre-training, MIM encourages models to aggregate adjacent pixels, thus increasing their ability to capture local features. 
\begin{figure}[t]
    \centering
    \includegraphics[width=0.35\textwidth]{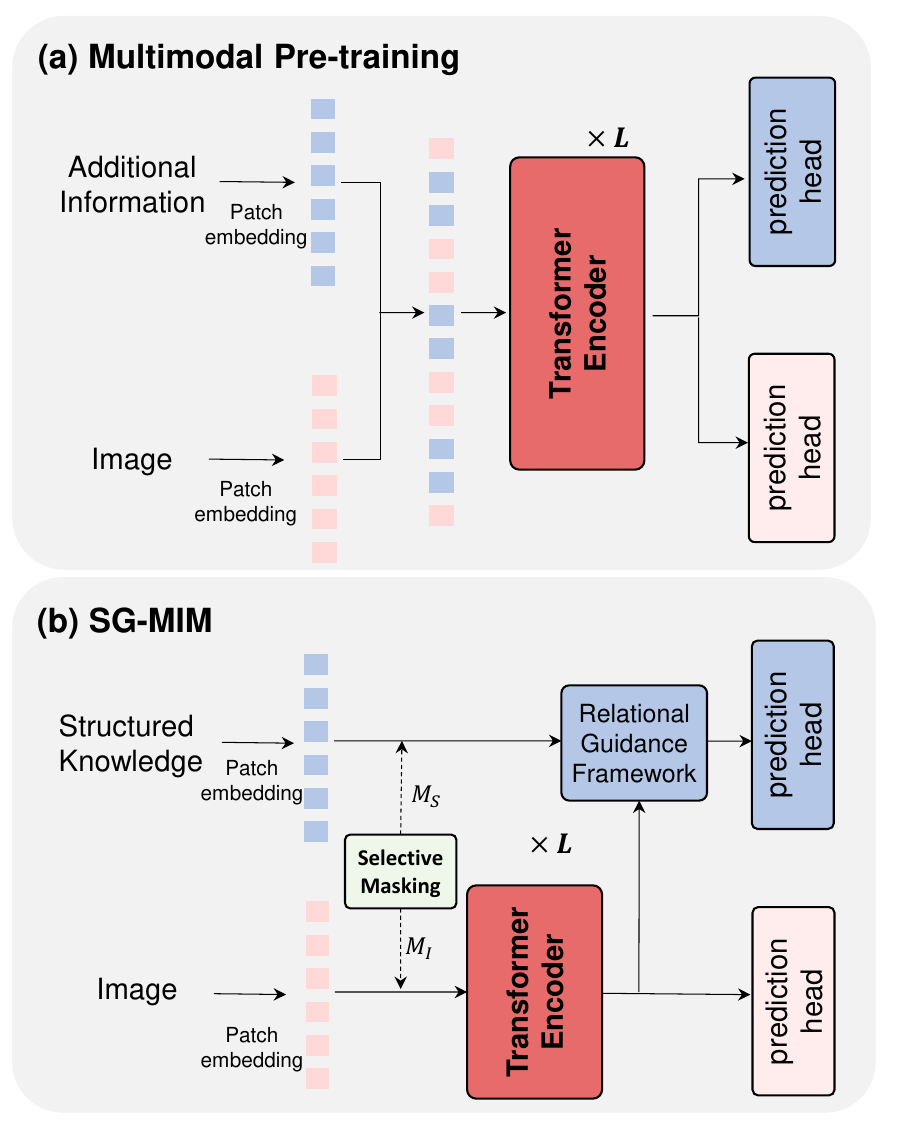} 
    \caption{{\bf Comparison with existing multimodal pre-training}: (a) Multimodal pre-training and (b) the proposed method (SG-MIM).  
    While a common form of multimodal pre-training method, \emph{e.g.},~\cite{bachmann2022multimae,weinzaepfel2022croco}, integrates both types of data directly into the Transformer encoder
    , SG-MIM uses a lighter relational guidance framework.
    }
    \label{fig:model_compare} 
\end{figure}

Yet, despite their impressive achievements, MIM models often fall short in generalizing effectively to dense prediction tasks such as monocular depth estimation~\cite{godard2019digging, choi2021adaptive, kim2022global} and semantic segmentation~\cite{long2015fully, chen2017deeplab, cho2024cat}. 
This is primarily due to the inherent lack of spatially structured information, such as relational cues between pixels, leading to a deficiency in essential data that must be effectively transferred during pre-training for downstream tasks.

 To address this issue, prior MIM models have investigated the integration of multiple modalities or additional images as input sources. These approaches typically employ architectures that naively combine an image with another modality or additional images, treating them as a unified input to the encoder, as illustrated in Figure~\ref{fig:model_compare}(a). For instance, CroCo~\cite{weinzaepfel2022croco} utilizes two images from different viewpoints of the same scene, while MultiMAE~\cite{bachmann2022multimae} integrates images with pseudo-depth and segmentation maps within the same architecture.

 However, this method of naively merging an image with supplementary data introduces several challenges. (1) First, it creates a discrepancy between the pre-training phase and the fine-tuning phase. During pre-training, the encoder processes multiple inputs, while in fine-tuning, it manages only a single image. This discrepancy restricts the model's ability to effectively leverage the diverse information from additional images and modalities. (2) Furthermore, the model is vulnerable to noise introduced by the supplementary data. Predicted depth and segmentation maps are often employed as additional data, yet directly feeding this unrefined input into the encoder at the pixel level inevitably degrades performance. (3) Finally, naively merging an image with supplementary data increases the information load on the encoder, requiring longer training times. For example, MultiMAE~\cite{bachmann2022multimae} demands double the pre-training epochs—1600 compared to the 800 used by models like MAE~\cite{he2022masked} and SimMIM~\cite{xie2022simmim}.

Building on the aforementioned challenges, we propose a strategically designed architecture that efficiently leverages additional structured data. Our \textbf{S}tructured knowledge \textbf{G}uided \textbf{M}asked \textbf{I}mage \textbf{M}odeling (\textbf{SG-MIM}) introduces an innovative architecture where the encoder indirectly learns spatially structured information via a lightweight relational guidance framework. By utilizing an independent feature extraction branch, the proposed framework efficiently encodes structured knowledge, effectively bridging the gap between pre-training and downstream tasks. Moreover, unlike existing approaches~\cite{bachmann2022multimae, weinzaepfel2022croco} that naively merge inputs at the pixel level, the proposed architecture separately encodes structured information and guides the main image encoder with a feature fusion module at the feature level. This feature-level guidance enhances robustness to noise by filtering out irrelevant information, allowing the model to focus on meaningful patterns and achieve a more comprehensive contextual understanding.



In addition to utilizing a well-designed framework that seamlessly integrates additional structured knowledge with image input, we propose a semantic selective masking approach that introduces heterogeneous masking between different input signals. Our semantic selective masking approach strategically chooses specific patches for masking by considering the balance of learning difficulty. This balanced approach enhances the effectiveness of the relational guidance framework, leading to more robust and efficient feature learning.

Our approach serves as a general solution that operates without the need for additional annotations, offering adaptability and efficiency across a wide range of tasks. Moreover, it facilitates the generation of fine-grained, texture-rich features that substantially boost performance in dense prediction tasks, as highlighted in the analysis presented in Figure~\ref{fig:log_frequency_result}. 
In experimental comparisons with other models, SG-MIM consistently demonstrated superior performance, particularly at lower epochs such as 100. Notably, our method achieved an RMSE of 2.04 on the KITTI validation dataset~\cite{geiger2013vision}, a \(\delta_1\) of 0.91 on the NYU-v2 validation dataset~\cite{silberman2012indoor}—where \(\delta_1\) represents the percentage of predicted pixels where the ratio between the predicted and true depth is within a threshold of 1.25— and an mIoU of 47.59 on the ADE20K dataset~\cite{zhou2017scene}, demonstrating superior performance in dense prediction tasks across various backbone models and epochs compared to existing MIM models.


 The contributions of our model can be summarized as follows:
\begin{itemize}
    \item We propose an efficient independent relational guidance framework to address the framework issues of existing models, which often cause discrepancies between pre-training models and downstream tasks and are vulnerable to noise in different modalities.
    \item We experimentally demonstrate that using a selective guidance masking strategy during pre-training effectively transfers structured knowledge to the image encoder by strategically focusing on patches that best balance the learning difficulty.
    \item Our method is an off-the-shelf approach with general applicability, capable of integrating into any backbone model without requiring additional annotations. Furthermore, our performance has been validated through diverse experiments on monocular depth estimation and semantic segmentation tasks across various backbones.
\end{itemize}

\section{Related Work}
\subsection{Masked Image Modeling (MIM)}
In the domain of computer vision, self-supervised learning has identified MIM~\cite{he2022masked,xie2022simmim} as playing a crucial role. Inspired by Masked Language Modeling from BERT~\cite{devlin2018bert}, MIM has demonstrated impressive performance in visual representation learning~\cite{grill2020bootstrap, chen2021exploring, ema1, ema2}. This approach involves learning visual representations by restoring pixels missing in images, a method that leverages the concept of learning through reconstruction. The success of MIM can be attributed to its ability to impart locality inductive bias~\cite{xie2023revealing} to the trained models, enabling the models to aggregate near pixels in the attention heads.

Currently, the MIM approach is exemplified by two main methodologies: MAE~\cite{he2022masked} and SimMIM~\cite{xie2022simmim}. MAE, utilizing ViT~\cite{dosovitskiy2020image} as its backbone, operates by inputting only visual image tokens into the encoder and integrating masked tokens just before entering the decoder, where the reconstruction occurs. On the other hand, SimMIM~\cite{xie2022simmim}, which can use ViT~\cite{dosovitskiy2020image} or Swin~\cite{liu2021swin} as its backbone, introduces both visual image tokens and masked tokens into the encoder, initiating reconstruction from the encoder stage itself. Consequently, the decoder in SimMIM is designed as a lightweight prediction head, distinguishing its architecture from MAE. This diversity in approaches underscores the adaptability and potential of MIM in advancing the field of visual representation learning.

\subsection{Variants of MIM}
Building on the success of MIM, numerous variations of its structure have been proposed to further extend its capabilities.
Croco~\cite{weinzaepfel2022croco} adopts a cross-view completion strategy, taking as inputs two images of the same scene from different views. Only one input image undergoes masking, and then a siamese encoder~\cite{dosovitskiy2020image} form is used to encode only the visible parts of the two images. Before entering the decoder, the masked tokens are combined with the encoded visible parts to reconstruct the masked tokens, facilitating learning from this integrated approach.
MultiMAE~\cite{bachmann2022multimae} utilizes methods for monocular depth estimation and semantic segmentation tasks to generate pseudo-depth and segmentation maps, which are then integrated with images as inputs. Distinct decoders for each modality are utilized to reconstruct the information, showcasing a comprehensive approach to multimodal visual representation learning. These variations on MIM illustrate the ongoing innovation in the field, aiming to exploit the full potential of self-supervised learning for enhancing visual understanding across a range of applications.



\section{Preliminary}
Masked Image Modeling (MIM) is a cornerstone technique in self-supervised learning for computer vision, where the model learns to reconstruct randomly masked portions of an input image. This process helps the model acquire general visual representations that are useful across various downstream tasks, such as classification, segmentation, and object detection. The reconstruction loss, typically calculated as L1 or L2 loss between the reconstructed and original pixels, guides the learning process. The loss is formulated as:
\[
L_{\text{rec}} = \frac{1}{N} \sum_{i=1}^{N} M_I(i) \cdot \left| I_p(i) - I(i) \right|
\]
where \(N\) denotes the total number of masked pixels, \(I_p(i)\) represents the reconstructed pixel values, and \(I(i)\) denotes the original pixel values. The mask indicator \(M_I(i)\) equals 1 if the \(i\)-th pixel is masked and 0 otherwise. The encoder, trained through MIM, is then used in downstream tasks, ensuring that the learned features are adaptable to various applications beyond image reconstruction.





\section{Method}
In this section, we introduce the SG-MIM framework, detailing its network architecture and presenting Fourier analysis to show how it enhances fine-grained feature generation and improves performance in dense prediction tasks.


\begin{figure*}[t]
    \centering
    \includegraphics[width=0.9\textwidth]{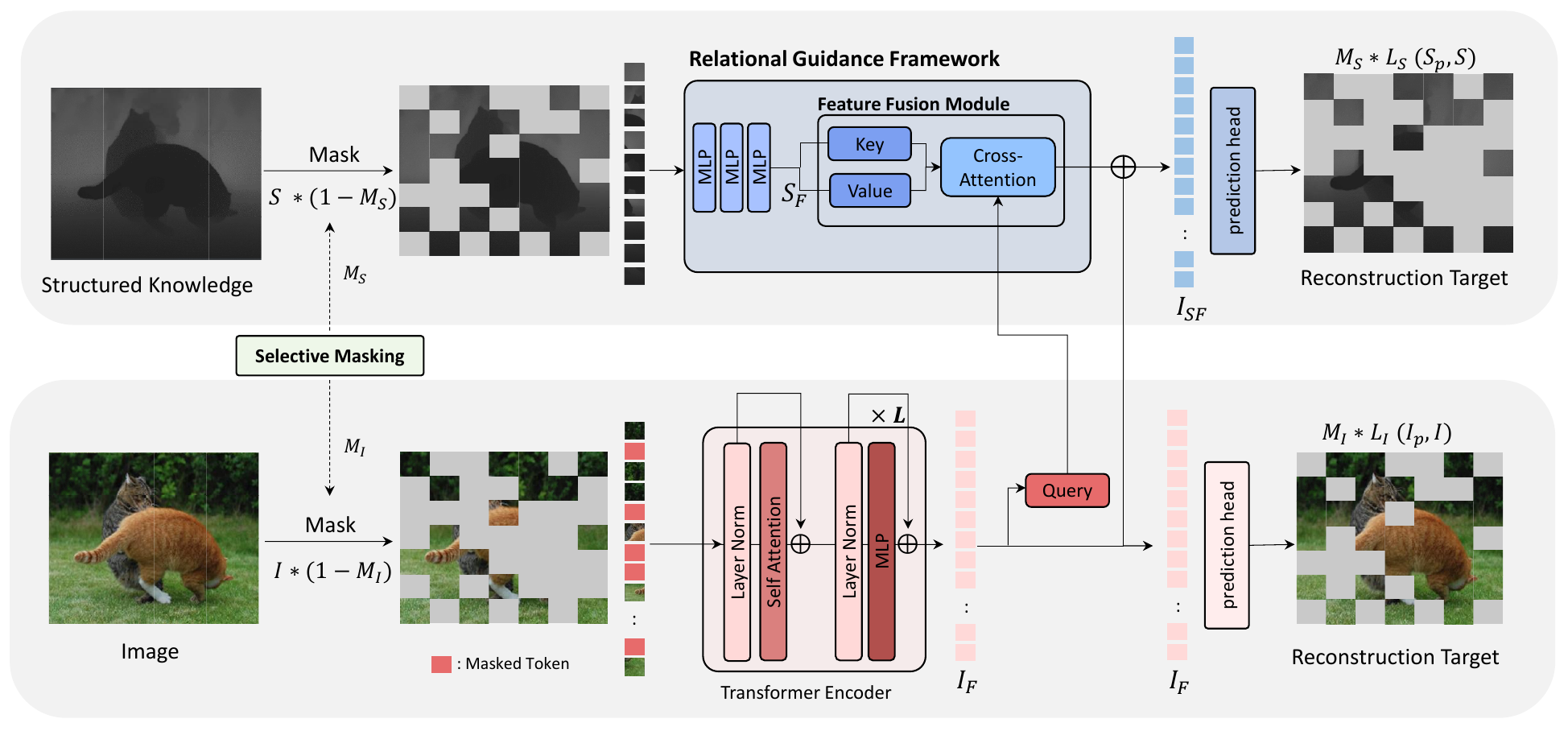} 
    \caption{{\bf Overview of the proposed SG-MIM.} Image and Structured knowledge map are masked in accordance with \(M_I\) and \(M_S\), respectively. The masked image, combined with masked tokens, enters the encoder (ViT~\cite{dosovitskiy2020image} or Swin transformer~\cite{liu2021swin}), resulting in the image latent representation \(I_F\). This proceeds to the image prediction head to predict the original image values for the missing patches. Simultaneously, \(I_F\) is transformed into a structured knowledge-guided image latent representation \(I_{SF}\) within the relational guidance framework, aided by \(S_F\) extracted through shallow MLP layers. This is then directed to the prediction head, arranged in parallel, to predict the structured information for the visible image patches. 
    Note that only the pre-trained Transformer encoder is used in the subsequent downstream tasks.} 
    \label{fig:network_arch} 
    \vspace{-0.2cm}
\end{figure*}

\subsection{Overview}
While the utilization of additional information during pre-training has been extensively studied, previous network architectures, as illustrated in Figure~\ref{fig:model_compare} (a), have typically relied on naive pixel-level integration. In contrast, SG-MIM leverages structured knowledge~\cite{ranftl2021vision} and  adopts an independent network architecture like Figure~\ref{fig:model_compare} (b), by incorporating a relational guidance framework that encodes structured information parallel to the traditional MIM architecture. The framework comprises key components: Selective Guidance Masking and Encoding, which strategically targets patches to adjust learning difficulty; the relational guidance framework, which independently encodes and fuses structured data; Prediction Head and Loss Function, which together optimize the model by combining image reconstruction and structured knowledge prediction to effectively balance general feature learning and structured information capture.

\paragraph{Selective Guidance Masking and Encoding}

The input image \(x \in \mathbb{R}^{H \times W \times C_i}\) is divided into patches \(x_p \in \mathbb{R}^{N \times (P^2 \cdot C_i)}\). Similarly, the structured knowledge map is also segmented into patches \(s_p \in \mathbb{R}^{N \times (P^2 \cdot C_s)}\). Here, \(N=HW/P^2\) denotes the number of divided patches having a resolution of $P\times P$. $C_i=3$ and $C_s=1$ represent channel size, respectively. These patches are then transformed into patch embeddings through their respective linear projections. The image patch embeddings follow the traditional MIM masking strategy, masking the majority of the patches (e.g., 60\%).

Meanwhile, the structured knowledge patch embeddings are masked using a semantic selective guidance masking strategy, which ensures that there is no overlap with the masked regions of the input image. By selectively utilizing structured knowledge patches, it ensures that only visible image patches contribute to the estimation of structured details. Furthermore, it prevents the model from trying to infer structured information from invisible image patches, which could unnecessarily complicate the learning process. This approach, grounded in a semantic perspective, focuses on selecting patches that enhance the synergy between structured knowledge and general representation learning.

This masking strategy can be mathematically expressed as follows. Let \(M_I\) and \(M_S\) represent the masking matrix for the image and structured knowledge patch embeddings, respectively. 
Both matrices are of dimension \(N \times 1\), consisting of elements in \(\{0, 1\}\), where 1 indicates an invisible (masked) patch and 0 otherwise.
Our selective masking strategy ensures that no overlap occurs in the masking of the image and structured knowledge map, formalized as \(M_{I,j} + M_{S,j} = 1\) each \(j\).

Following this masking strategy, the visible image patch embeddings, along with learnable masked tokens, are input into the transformer encoder~\cite{dosovitskiy2020image,liu2021swin} to create an image latent representation \(I_F\), while the visible structured knowledge patch embeddings are processed by the relational guidance framework to guide the model with structured knowledge. An ablation study in Table~\ref{tab:combined_masking_loss} investigates the effects of different masking strategies.

\paragraph{Relational Guidance Framework}

The relational guidance framework is a lightweight module designed to encode structured knowledge using MLP layers, specifically aligned with the hierarchical image encoder. By maintaining an independent encoding structure, this module effectively avoids discrepancies with downstream tasks and mitigates the increased learning burden on the encoder.

Our framework receives inputs from the structured knowledge patch embeddings and image latent representations, \(I_F\). It can be divided into two main components: feature extraction comprising shallow MLP layers, which generates structured knowledge features \(S_F\), and a feature fusion module that fuses \(S_F\) with the image latent representation \(I_F\). 
This shallow feature extraction demonstrates greater efficiency in terms of training complexity (refer to Table~\ref{tab:encoder_efficiency}). 

Given that structured knowledge contains simpler information compared to images, our method attempts to represent the structured knowledge using shallow MLP layers instead of the computational heavy Transformer encoder~\cite{liu2021swin}. This approach mirrors the methodology adopted by PointNet~\cite{qi2017pointnet}, which utilizes MLPs to derive point features from 3D point clouds, highlighting the efficiency of MLPs in processing 3D geometric data.  

Also, the feature fusion module facilitates the learning of relationships between the two modalities, enabling the generation of a structured-guided image latent representation \(I_{SF}\) for the visible parts of the image. This is achieved with the help of patches corresponding to areas that are visible in the structured knowledge map (but invisible in the image). The feature fusion module can be implemented as a residual connection structure of a multi-head cross-attention layer with the image latent representation \(I_F\) (query) and the structured feature \(S_F\) (key and value), as shown in Figure~\ref{fig:network_arch}. 




Within a feature fusion module, the query, key, and value projections for each head $i$ are defined as:
\begin{align*}
Q_i = W^Q_i I_F, \quad\quad K_i = W^K_i S_F, \quad\quad V_i = W^V_i S_F,
\end{align*}
where $W^Q_i$, $W^K_i$, and $W^V_i$ are learned weights. The multi-head cross-attention mechanism enriches the image features by integrating these projections:

\begin{equation*}
I_{SF} = \text{Concat}(\text{head}_1, ..., \text{head}_h)W^O + I_F,
\end{equation*}
\begin{equation*}
\text{where} \quad \text{head}_i = \text{attention}(Q_i, K_i, V_i),
\end{equation*}

Here, $I_{SF}$ represents the structured-guided image latent representation, enhanced through multi-head cross attention, combining the outputs from all heads.

\paragraph{Prediction Head and Loss Function}
In our SG-MIM model, the image latent representation \(I_F\), processed by the Transformer encoder~\cite{dosovitskiy2020image,liu2021swin}, is fed into a lightweight, one-layer prediction head similar to SimMIM~\cite{xie2022simmim}. The \emph{image reconstruction} loss \(L_{\text{I}} = \frac{1}{N} \sum_{i=1}^{N} M_I(i) \cdot \left| I_p(i) - I(i) \right|\) is calculated using L1 loss between the reconstructed pixels \( I_p(i)\) and the target image pixels \( I(i)\), where \(N\) is the total number of masked pixels, and \(M_I(i)\) is derived from traditional MIM masking.

In parallel, the structured-guided latent representation is processed through a separate prediction head designed for handling structured information, resulting in the \emph{structured knowledge prediction} loss \(L_{\text{S}} = \frac{1}{N} \sum_{i=1}^{N} M_S(i) \cdot \left| S_p(i) - S(i) \right|\), which also uses L1 loss to compare predicted structured knowledge \(S_p(i)\) with target values \(S(i)\).

The total loss function combines these two losses, optimizing the model to learn both general and structured features effectively:
\[
L = \lambda_{\text{I}} L_{\text{I}} + \lambda_{\text{S}} L_{\text{S}},
\]
where \(\lambda_{\text{I}}\) and \(\lambda_{\text{S}}\) balance the contributions of image reconstruction and structured knowledge prediction losses. In our experiments, both weights are set to 1, with an ablation study presented in Table~\ref{tab:combined_masking_loss}.

\subsection{Fourier Analysis of Feature Maps}
\begin{figure}[t]
    \centering
    \includegraphics[width=0.4\textwidth]{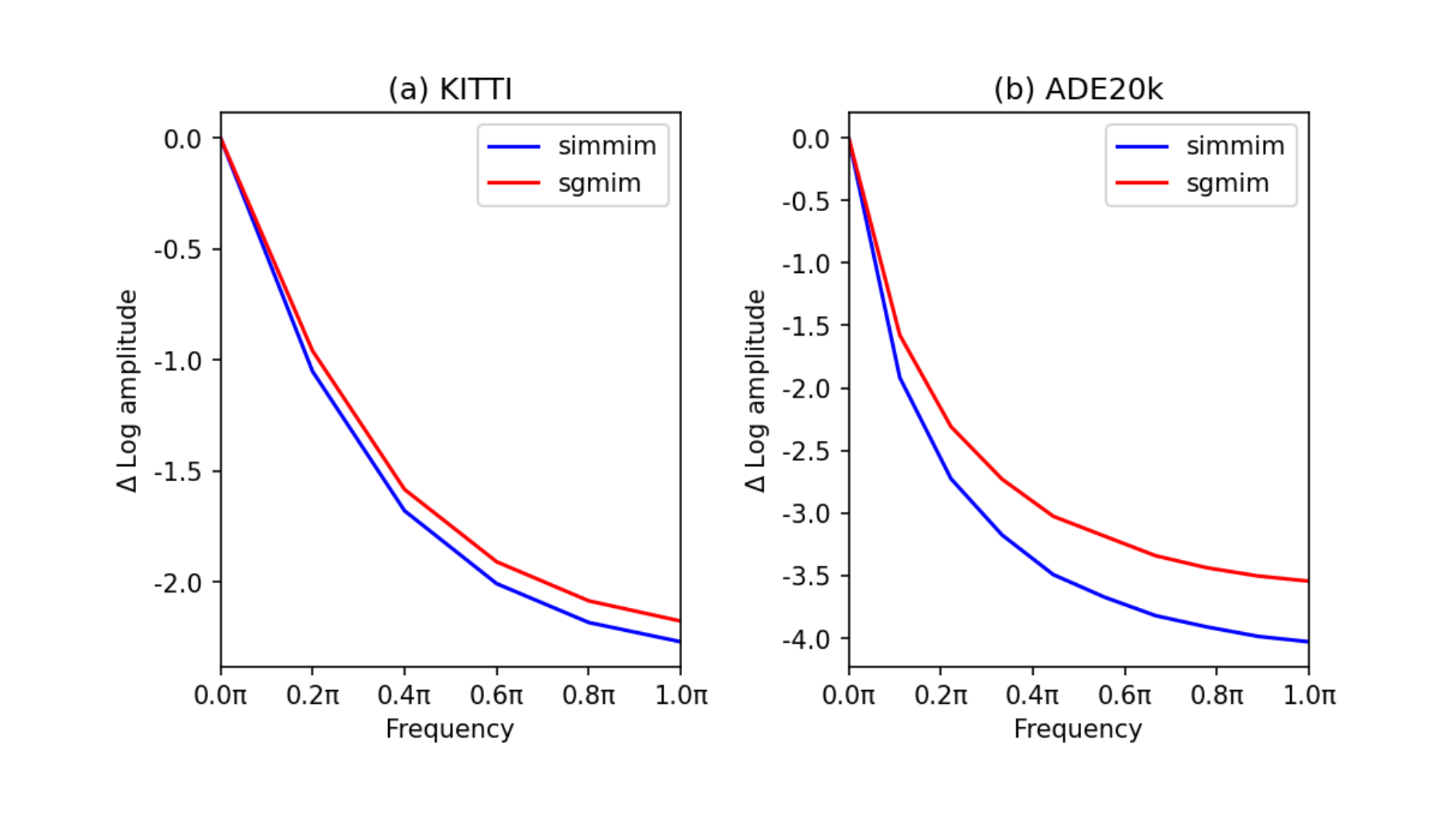} 
    \caption{{\bf Relative log amplitudes of Fourier transformed feature maps in Dense Prediction}: We present the comparison of relative log amplitudes in Fourier transformed feature maps between SG-MIM and SimMIM~\cite{xie2022simmim} for dense prediction tasks. Panel (a) illustrates the feature maps for depth estimation on the KITTI~\cite{geiger2013vision} validation dataset, and panel (b) displays the feature maps for segmentation on the ADE20K~\cite{zhou2017scene} validation dataset. }
    \label{fig:log_frequency_result} 
    \vspace{-0.2cm}
\end{figure}



We conducted a visualization analysis using Fourier analysis to compare the features produced by SG-MIM and SimMIM. Specifically, the $\Delta$Log amplitude is calculated as the difference between the log amplitude at normalized frequency $0.0\pi$ (center) and at $1.0\pi$ (boundary). For better visualization, we only provide the half-diagonal components of the two-dimensional Fourier-transformed feature map. Figure~\ref{fig:log_frequency_result} shows that SG-MIM effectively captures high-frequency signals, which facilitates the generation of more detailed features with rich edges and textures. This capability is particularly beneficial for dense prediction tasks, where such fine-grained textural information is crucial for improved performance. The analysis was conducted on the KITTI dataset for depth estimation and the ADE20K dataset for semantic segmentation, demonstrating SG-MIM's superior ability to capture essential high-frequency details across different types of dense prediction tasks.

\begin{table*}[t]
\tiny
\centering
\resizebox{\textwidth}{!}{%
\begin{tabular}{cccccc}
\hline
\textbf{Methods}  & \textbf{Task} & \textbf{Backbone} & \textbf{Epoch} & \textbf{Data}   & \textbf{RMSE} $\downarrow$ \\ \hline
Multi-MAE~\cite{bachmann2022multimae}   & RGB+D+S       & ViT-B             & 1600           & ImageNet        & 2.36          \\
Croco~\cite{weinzaepfel2022croco} &Cross View RGB           & ViT-B             & 400            & Habitat  & 2.44          \\
MAE~\cite{he2022masked}       & RGB           & ViT-B             & 800            & ImageNet        & 2.26          \\
SimMIM~\cite{xie2022simmim}   & RGB           & ViT-B             & 800            & ImageNet        & 2.23             \\
SimMIM~\cite{xie2022simmim}    & RGB           & Swin-B            & 100            & ImageNet        & 2.49          \\
SimMIM~\cite{xie2022simmim}    & RGB           & Swin-B            & 800            & ImageNet        & 2.23          \\ \hline
SG-MIM            & RGB+D         & Swin-B            & 100            & ImageNet        & \textbf{2.29}          \\
SG-MIM            & RGB+D         & ViT-B             & 800            & ImageNet        & \textbf{2.20}          \\
SG-MIM            & RGB+D         & Swin-B            & 800            & ImageNet        & \textbf{2.19}          \\ \hline
\end{tabular}%
}

\caption{{\bf Monocular Depth Estimation on KITTI Val Dataset~\cite{geiger2013vision}.} A comparison of our model's performance with existing MIM models~\cite{bachmann2022multimae,weinzaepfel2022croco,he2022masked,xie2022simmim} using RMSE as the metric. The task column indicates the data being restored (\(D\) for depth, \(S\) for segmentation), comparing the ViT-Base~\cite{dosovitskiy2020image} (\(224 \times 224\)) and Swin-Base~\cite{liu2021swin} backbones (\(192 \times 192\)) across different pre-training epochs. Unlike other models, Croco~\cite{weinzaepfel2022croco} employs the Habitat dataset~\cite{ramakrishnan2021habitat}, which contains a larger number of images than ImageNet~\cite{deng2009imagenet}.}
\label{tab:depth_kitti_tab}
\end{table*}


\begin{table}[]
\centering
\begin{tabular}{cccc}
\hline
\textbf{Methods}                                   & \textbf{Backbone}                                  & \textbf{Epoch} & \textbf{\begin{tabular}[c]{@{}c@{}}\textbf{RMSE} $\downarrow$\end{tabular}} \\ \hline
SimMIM                                             & Swinv2-B                                           & 100            & 2.30                                                            \\
SimMIM                                             & Swinv2-B                                           & 800            & 2.06                                                            \\ \hline
SG-MIM                                             & Swinv2-B                                           & 100            & 2.19                                                            \\
SG-MIM                                             & Swinv2-B                                           & 800            & \textbf{2.04}                                                            \\ \hline
\multicolumn{2}{c|}{\multirow{2}{*}{\begin{tabular}[c]{@{}c@{}}Representative \\ Methods\end{tabular}}} & BinsFormer     & 2.09                                                            \\ \cline{3-4} 
\multicolumn{2}{c|}{}                                                                                   & iDisc          & 2.06                                                            \\ \hline
\end{tabular}
\caption{{\bf Monocular Depth Estimation compared to representative methods.} This table compares the RMSE performance on the KITTI validation dataset~\cite{geiger2013vision}, using the metric to evaluate SG-MIM and SimMIM with a SwinV2-Base~\cite{liu2022swin} backbone, against established monocular depth estimation models such as BinsFormer~\cite{li2024binsformer} and iDisc~\cite{piccinelli2023idisc}.}
\label{tab:depth_kitti_representative}
\end{table}

\subsection{Implementation Details}
In our pre-training phase, we conducted experiments leveraging Swin-Base~\cite{liu2021swin}, Swinv2-Base~\cite{liu2022swin}, and ViT-Base~\cite{dosovitskiy2020image}. 
The default input sizes for Swin Transformer and ViT are set to \(192 \times 192\) and \(224 \times 224\), respectively, with a uniform image masking ratio of 0.6 across all tests. The structured knowledge is generated using a DPT-Hybrid~\cite{ranftl2021vision} trained on the OmniData~\cite{eftekhar2021omnidata}. 
Training is conducted with a batch size of 1024 on 8 GPUs of NVIDIA RTX 6000 Ada. 
Additional experiments and implementation details are available in the Supplementary material.

\section{Experiments}
In this section, we conducted a series of experiments to compare the fine-tuning performance of our model against existing pre-training models~\cite{he2022masked,xie2022simmim,bachmann2022multimae,weinzaepfel2022croco} across a variety of tasks, including monocular depth estimation, semantic segmentation. The experimental setup is organized as follows: we begin with monocular depth estimation experiments, followed by semantic segmentation, and conclude with model efficiency and an ablation study. 

\subsection{Downstream Task: Monocular Depth Estimation}

\noindent \textbf{Data and Setup}
For the monocular depth estimation experiments, we utilized the standard dataset splits for both the KITTI~\cite{geiger2013vision} and NYU-v2~\cite{silberman2012indoor} benchmarks. 
For the KITTI dataset, inspired by GLPDepth~\cite{kim2022global}, we appended a simple depth estimation head consisting of deconvolution layers to the encoder~\cite{dosovitskiy2020image, liu2021swin}. We adopted RMSE as the evaluation metric. 

For the NYU-v2 dataset, we employed the DPT~\cite{ranftl2021vision} with encoder~\cite{dosovitskiy2020image}, evaluating performance with the metric \(\delta_1\)~\cite{doersch2017multi}, e.g., \(\left( \frac{d_{\text{{gt}}}}{d_{\text{{p}}}}, \frac{d_{\text{{p}}}}{d_{\text{{gt}}}} \right)\), which represents the percentage of pixels where the relative depth error is less than 1.25. 
Here, \(d_{\text{{p}}}\) and \(d_{\text{{gt}}}\) denote the predicted depth and ground truth depth, respectively.\\

\noindent \textbf{Result} In the performance comparison across downstream models, SG-MIM consistently demonstrates superior results compared to existing MIM models~\cite{bachmann2022multimae, weinzaepfel2022croco, he2022masked, xie2022simmim}. As shown in Table~\ref{tab:depth_kitti_tab}, SG-MIM improves upon the baseline model, SimMIM~\cite{xie2022simmim}, across all configurations, including both ViT-Base and Swin-Base backbones, at 100 and 800 epochs (noting that lower RMSE indicates better performance). Additionally, compared to other MIM models, such as MultiMAE~\cite{bachmann2022multimae}, which involves a more complex reconstruction task (RGB+D+S), SG-MIM outperforms these models when utilizing the same ViT-Base backbone. Additionally, even though Croco~\cite{weinzaepfel2022croco} uses a larger dataset, specifically the Habitat dataset~\cite{ramakrishnan2021habitat}, which includes 1,821,391 synthetic image cross-view pairs, SG-MIM still achieves better performance. 

As shown in Table~\ref{tab:depth_kitti_representative}, we evaluated our model not only against other MIM-based models but also against models specifically designed for monocular depth estimation. In this comparison, both SimMIM and SG-MIM were pre-trained using the Swinv2-Base backbone, with the trained encoder weights transferred to the GLPDepth model for performance evaluation. For representative methods, we included state-of-the-art models such as BinsFormer~\cite{li2024binsformer} and iDisc~\cite{piccinelli2023idisc}. Compared to SimMIM using the same downstream model, SG-MIM showed a significant performance improvement at 100 epochs and a slight improvement at 800 epochs. Furthermore, SG-MIM demonstrated comparable or superior performance when compared to state-of-the-art models.

In Table~\ref{tab:depth_NYU_tab}, where the downstream model is implemented using DPT based on the Vit-Base backbone. Similar to Table~\ref{tab:depth_kitti_tab}, SG-MIM demonstrates superior performance in the  \(\delta_1\) metric. Interestingly, contrary to Table~\ref{tab:depth_kitti_tab}, Croco~\cite{weinzaepfel2022croco} exhibits higher performance among other MIM pre-training models, achieving the same \(\delta_1\) score as SG-MIM, while MAE~\cite{he2022masked} shows the lowest performance. However, it should be noted that Croco has been pre-trained with a larger quantity of images~\cite{ramakrishnan2021habitat} than other models. 

\begin{table}[t]
\centering
\resizebox{\columnwidth}{!}{%
\begin{tabular}{ccccc}
\hline
\textbf{Methods}  & \textbf{Task} & \textbf{Epoch} & \textbf{Data}   & \textbf{$\delta_1$} $\uparrow$ \\ \hline
Multi-MAE& RGB+D+S       & 1600           & ImageNet        & 0.88          \\
Croco & Cross View RGB           & 400            & Habitat& 0.91          \\
MAE         & RGB           & 800            & ImageNet        & 0.87          \\
SimMIM    & RGB           & 800            & ImageNet        & 0.89             \\ \hline
SG-MIM            & RGB+D         & 800            & ImageNet        & \textbf{0.91}          \\ \hline
\end{tabular}%
}
\caption{{\bf Monocular Depth Estimation on NYU-v2 Dataset.} For the NYU-v2 dataset, the downstream model is DPT~\cite{ranftl2021vision}, and all experiments are conducted with the Vit-base~\cite{dosovitskiy2020image} backbone, using  \(\delta_1\) as the metric. Unlike other models, Croco~\cite{weinzaepfel2022croco} employs the Habitat dataset~\cite{ramakrishnan2021habitat}, which contains a larger number of images than ImageNet~\cite{deng2009imagenet}.}
\label{tab:depth_NYU_tab}
\end{table}

\subsection{Downstream Task: Semantic Segmentation}
\noindent \textbf{Data and Setup}
We conducted semantic segmentation experiments on the ADE20K~\cite{zhou2017scene} dataset. The UperNet framework~\cite{xiao2018unified} served as the downstream model, with pre-trained weights loaded into the encoder for finetuning. The performance was evaluated using the mIoU metric, and further details of the experimental setup and results can be found in the Supplementary material.\\

\noindent \textbf{Result}
As shown in Table~\ref{tab:ade20k_seg}, we validated the performance of our model, SG-MIM, on the semantic segmentation task using the ADE20K validation dataset under the same conditions as SimMIM with the SwinV2-Base backbone. Our results demonstrate that SG-MIM achieved an approximately 0.5 higher mIoU score than SimMIM. Additionally, it consistently outperformed other models, such as MultiMAE.

\begin{table}[]
\centering
\begin{tabular}{llll}
\hline
\textbf{Methods} & \textbf{Backbone} & \textbf{Epoch} & \textbf{mIoU$\uparrow$} \\ \hline
MoCov3           & ViT-B             & 1600            & 43.7          \\
DINO             & ViT-B             & 1600            & 44.6          \\
MAE              & ViT-B             & 1600           &   46.2        \\
Multi-MAE        & ViT-B             & 1600           & 46.2          \\ \hline
SimMIM           & Swinv2-Base       & 800            &   47.05            \\
SG-MIM           & Swinv2-Base       & 800            & \textbf{47.59}          \\ \hline
\end{tabular}
\caption{{\bf Semantic Segmentation on ADE20K Dataset.} This table presents the results of semantic segmentation on the ADE20K~\cite{zhou2017scene} dataset, using the UperNet~\cite{xiao2018unified} framework as the downstream model and mIoU as the evaluation metric. The mIoU scores for MoCov3~\cite{chen2021empirical}, DINO~\cite{caron2021emerging}, MAE~\cite{he2022masked}, and Multi-MAE~\cite{bachmann2022multimae}, all using ViT as the backbone, are sourced from the MultiMAE.}
\label{tab:ade20k_seg}
\end{table}

\subsection{Model Efficiency}
In Table~\ref{tab:encoder_efficiency}, we examine the efficiency of SG-MIM based on different feature extraction architectures in the relational guidance framework—MLP layers, Transformer~\cite{liu2021swin}, and Siamese Transformer~\cite{liu2021swin}—and their performance in monocular depth estimation on the KITTI dataset~\cite{geiger2013vision}. The Transformer architecture operates independently from the image encoder, while the Siamese Transformer shares weights with the image encoder, indicating a unified processing approach. SG-MIM with MLP-based encoding excels in both training efficiency and RMSE performance. Interestingly, Transformer-based models show lower performance, likely due to their higher capacity requiring longer training times than the 800 epochs used in our experiments. This highlights the suitability of MLPs for capturing structured features efficiently.

\begin{table}[t]
\centering
\resizebox{\columnwidth}{!}{%
\begin{tabular}{cccc}
\hline
\textbf{Feature extraction} & \textbf{\begin{tabular}[c]{@{}c@{}}Training \\ Time\end{tabular}} & \textbf{\begin{tabular}[c]{@{}c@{}}Memory\\ Consumption\end{tabular}} & \textbf{\begin{tabular}[c]{@{}c@{}}RMSE$\downarrow$\end{tabular}} \\ \hline
MLP                    & \textbf{8m 55s}                                                   & \textbf{24.6GB}                                                       & \textbf{2.29}                                                   \\
Transformer            & 15m 24s                                                           & 39.6GB                                                                & 2.37                                                            \\
Siamese Transformer    & 13m 34s                                                           & 39.0GB                                                                & 2.67                                                            \\ \hline
\end{tabular}%
}
\caption{{\bf Efficiency Analysis by feature extraction in the relational guidance framework .} This table compares the training time per epoch (``m" represents minutes, and ``s" represents seconds.), memory consumption per GPU, and performance on the KITTI dataset~\cite{geiger2013vision} according to different structures of the feature extraction, with all Image Encoders utilizing a transformer architecture~\cite{liu2021swin, chen2021pre, ranftl2021vision, lee2022knn, lee2023cross}.}
\label{tab:encoder_efficiency}
\vspace{-0.2cm}
\end{table}

\subsection{Ablation study}
All ablation studies are conducted on the KITTI dataset~\cite{geiger2013vision}, focusing on the monocular depth estimation using the Swin-Base as a backbone at 100 epochs.\\

\noindent \textbf{Masking Strategy and Ratio}
The study starts with traditional random masking, applied at a 0.6 ratio to both images and structured information. This can complicate the task of estimating structured information for invisible image patches, leading to poorer performance compared to ours, as shown in Table~\ref{tab:combined_masking_loss}. However, our selective masking strategy avoids overlap between masked regions in the image and structured information, allowing the model to focus on visible patches and effectively estimate structured details, achieving an RMSE of 2.29, as shown in Table~\ref{tab:combined_masking_loss}. We also experimented with adjusting the masking ratio from the default 0.6 to 0.5 and 0.7. Our results indicate that the 0.6 ratio achieves the best performance, yielding an RMSE of 2.29

\begin{table}[t]
\centering
\scriptsize
\renewcommand{\arraystretch}{0.8} 
\resizebox{\columnwidth}{!}{%
\begin{tabular}{cc|cc}
\hline
\multicolumn{2}{c|}{\textbf{Masking Strategy}} & \multicolumn{2}{c}{\textbf{Loss Weights}} \\ 
\textbf{Strategy} & \textbf{RMSE $\downarrow$} & \textbf{\(\lambda_{\text{rec}} / \lambda_{\text{dep}}\)} & \textbf{RMSE $\downarrow$} \\
\hline
Random (0.6) & 2.36 & 1/1 & \textbf{2.29} \\
Ours (0.6) & \textbf{2.29} & 1/0.1 & 2.32 \\
Ours (0.5) & 2.36 & 1/0.01 & 2.49 \\
Ours (0.7) & 2.39 & 1/0 & 2.49 \\
\hline
\end{tabular}
}
\caption{{\bf Comparative Analysis of Masking Strategies and Loss Weight Ratios.} This table compares RMSE performance across different masking strategies and ratios on the left and the impact of varying the \(\lambda_{\text{I}} / \lambda_{\text{S}}\) ratio for loss weights on the right for the KITTI validation dataset~\cite{geiger2013vision}.}
\label{tab:combined_masking_loss}
\vspace{-0.2cm}
\end{table}
\noindent \textbf{Loss Weights}
In Table~\ref{tab:combined_masking_loss}, experiments show that a balanced 1/1 ratio between image reconstruction and structured knowledge prediction losses yields the best RMSE of 2.29. Reducing the weight of the structured knowledge loss results in a progressive decline in performance, highlighting the importance of the relational guidance framework for optimal monocular depth estimation.

\section{Conclusions}
In conclusion, SG-MIM enhances Masked Image Modeling by effectively integrating structured knowledge into the pre-training process through a lightweight relational guidance framework. This enables efficient encoding of spatially structured information, reduces noise, and better aligns pre-training with downstream tasks. Additionally, the selective masking strategy manages learning difficulty by focusing on visible image regions, ensuring the model doesn't strain to predict structured details from areas lacking information. This efficient and balanced approach enables the model to generate fine-grained features, leading to improved performance in dense prediction tasks, particularly in depth estimation and semantic segmentation, where SG-MIM outperforms existing methods.\\

\noindent \textbf{Limitations} While SG-MIM effectively integrates structured data into the pre-training process, it is still inherently limited by the 2D nature of traditional MIM frameworks, which focus on reconstruction and prediction within a 2D plane.
Future work will address the limitation by extending the MIM framework to incorporate 3D point cloud data, enabling richer 3D perception and understanding tasks.

\clearpage

\bibliography{aaai25}
\newpage
\onecolumn

\end{document}